%% file: main.tex
\pdfoutput=1

\documentclass[11pt]{article}

\usepackage{acl}
\usepackage{times}
\usepackage{latexsym}
\usepackage{soul}
\usepackage[T1]{fontenc}
\usepackage[utf8]{inputenc}
\usepackage{scrextend}
\usepackage{microtype}
\usepackage[ruled,vlined]{algorithm2e}
\usepackage{listings}
\usepackage{authblk}
\usepackage{amsmath}
\usepackage{amssymb}
\usepackage{graphicx}
\usepackage{stackengine}
\usepackage{xcolor}
\usepackage{lipsum}
\usepackage{enumitem}
\usepackage{booktabs}
\usepackage{multirow}
\usepackage{hyperref}
\usepackage{longfbox} 

\definecolor{patterncolor2}{HTML}{d5eef2}
\definecolor{bred}{HTML}{f4cccc}
\definecolor{bgreen}{HTML}{d9ead3}
\definecolor{byellow}{HTML}{fff2cc}
\makeatletter
\newdimen\@tempdimd
\makeatother
\newfboxstyle{patternparam}{padding=1pt, margin-bottom=0pt, margin-top=0pt, border-radius=4pt, border-style=none, height=9pt, background-color=patterncolor2}

\def\green#1{\lfbox[patternparam, background-color=bgreen]{#1}}
\def\yellow#1{\lfbox[patternparam, background-color=byellow]{#1}}
\def\greenund#1{\green{\ul{#1}}}

\def\toolname{\mbox{GlotScript-T}\xspace}
\def\dataname{\mbox{GlotScript-R}\xspace}
\def\genericname{\mbox{GlotScript}\xspace}
\def\udhr{UDHR\xspace}
\def\numberlanguages{7,000\xspace}
\def\MAIN{CORE\xspace}

\title{\genericname:
A Resource and Tool for \\ Low Resource Writing System Identification}

\author[*$\diamond$]{Amir Hossein Kargaran}
\author[$\dag$]{François Yvon}
\author[*$\diamond$]{Hinrich Sch\"utze}

\affil[*]{Center for Information and Language Processing, LMU Munich, Germany}
\affil[$\diamond$]{Munich Center for Machine Learning (MCML), Germany}
\affil[$\dag$]{Sorbonne Université, CNRS, ISIR, France \protect\\
\texttt{amir@cis.lmu.de}}

\def\figref#1{Figure~\ref{fig:#1}}
\def\figlabel#1{\label{fig:#1}\label{p:#1}}

\def\tabref#1{Table~\ref{tab:#1}}

\def\tablabel#1{\label{tab:#1}\label{p:#1}}
\def\eqref#1{Eq.~\ref{eqn:#1}}

\def\seclabel#1{\label{sec:#1}\label{p:#1}}
\def\secref#1{\S\ref{sec:#1}}

\def\ucr{\scalebox{1}{\stackinset{c}{}{c}{-.1pt}{%
  \textcolor{white}{\sffamily\bfseries\small ?}}{%
  \rotatebox{45}{$\blacksquare$}}}}

\begin{document}
\maketitle
\begin{abstract}
We present \genericname,
an open resource and tool for low resource
writing system identification. 
\dataname is a resource that provides the attested writing systems for more than \numberlanguages languages.
It is compiled by aggregating information from
existing writing system resources.  
\toolname is a  writing system identification tool that covers all
161 Unicode 15.0 scripts.
For an input text,
it returns its script distribution where scripts are
identified by  ISO 15924 codes.
We also present two use cases for \genericname.
First, we demonstrate that \genericname
can help cleaning multilingual corpora such as mC4
and OSCAR. Second, we analyze the
tokenization  of a number of language models such as GPT-4
using \genericname and
provide insights on the coverage of low resource
scripts and languages by each language model. We hope that
\genericname will
become a useful resource for work on
low resource languages in the NLP community. \dataname and \toolname are available at  \url{https://github.com/cisnlp/GlotScript}.

\end{abstract}

\section{Introduction}
We are interested in automatically identifying the writing
system or script a given text is written in.  We will refer
to this automatic identification of scripts as \emph{script
identification}.  When doing research on and developing
technology for low resource languages, script identification
is useful. For example, when compiling a corpus for a
low resource language, script identification can serve as
part of quality control: texts written in scripts not used
for the language can be excluded. Similarly, when training
the tokenizer of a language model for low resource
languages, an analysis of the learned token vocabulary
allows developers to see how well a script is represented, an
indication of how well languages written in that script are
represented.

In such low resource scenarios, language identification is
an alternative to script identification: language identification can also be used
for quality control and for the analysis of language model
vocabularies. However, language identification for
low resource languages is prone to high error rates~\cite{kargaran2023glotlid, kreutzer2022quality,caswell2020language}.
Many low resource languages are poorly identified by existing tools,
due to data scarcity and high variability in orthography,
genres and
domains. By contrast, script
identification can be performed with a much higher accuracy and it is
therefore a useful functionality in the abscence of reliable
language identification for many low resource languages.

In this paper,
we present \genericname,
a resource and tool for low resource
identification of writing systems, i.e., low resource script identification.

Our contributions are as follows.
(i) We compile and organize \dataname, a comprehensive resource for script
identification, associating attested writing systems with language varieties.
We make this resource available to the community.
(ii) We publish \toolname, a tool for script identification
which covers all 161~scripts in Unicode 15.0. It
computes the script distribution for any input text. Scripts
are identified by their ISO~15924 codes.
To the best of our knowledge, no such tool is currently available.
(iii) We demonstrate the benefits of \toolname and \dataname for corpus cleaning,
and show that the quality of existing low resource corpora can be improved using script identification.
(iv) We analyze the tokenization of large language models
(LLMs) -- including
GPT-4, Falcon and Llama2 -- using \toolname. This analysis gives valuable insights
regarding LLM coverage (or lack of coverage) of low resource languages.

\section{Background and related work}\seclabel{background}

\subsection{Script identification} 

The Stops library~\cite{andrews-etal-2022-stopes}, part of the NLLB project~\cite{nllbteam2022language}, is capable of detecting the script of a given text in 38 scripts based on ISO 15924. It uses the Unicode blocks defined for each script.

\citet{Acs2019} gathered Unicode data block ranges and
mapped them to 18 macro Unicodes. For instance, they
categorized ranges like "Basic Latin", "C1 Controls and
Latin-1 Supplement", "Latin Extended Additional", "Latin
Extended-A", and "Latin Extended-B" into the Latin
script. These ranges were then employed with regular
expressions to identify the script of an input text.

These methods do not cover all of the
161 scripts that
Unicode 15.0 defines. Additionally, since
they use entire
blocks,\footnote{https://unicode.org/Public/15.0.0/ucd/Blocks.txt}
the result may not be entirely accurate. For example, the
range from U+0000 to U+007F is part of the "Basic Latin"
block. However, within this range, there are some common
characters that do not belong to a specific script and can
be used universally, such as the left square bracket
(U+005B). Compared to blocks, using a more granular approach,
in particular a per-character approach,\footnote{https://unicode.org/Public/15.0.0/ucd/Scripts.txt}
can be beneficial.

Python has the built-in library unicodedata.\footnote{https://docs.python.org/3/library/unicodedata.html} It allows
working with the Unicode Database. The command
\texttt{unicodedata.name(char)} can be used to obtain the name of a
character. This command only works for single
characters. However, the character's name does not always
include the name of its script.
Even if the name of the character contains information about
its script, there is no direct and consistent correspondence
of that information to  the codes of the  ISO~15924
standard.

\subsection{Language resources}

Many existing resources that compile
information about the world's languages, such as
Ethnologue~\cite{ethnologue},\footnote{The free access version.}
Glottolog~\cite{hammarstrom2023glottolog} and WALS (World
Atlas of Language Structures)~\cite{dryerwals}
do not contain information about writing systems.

Our work is most closely related to the work of
\citet{van-esch-etal-2022-writing}. Apart from the fact that
\citet{van-esch-etal-2022-writing} do not provide
script identification software
for use cases such as corpus cleaning, our approach differs in
methodology.
\citet{van-esch-etal-2022-writing} also aim to establish an
extensive metadata repository including writing systems and
speaker details. They cover  more than 2,800 languages.
But
their methodology
heavily focuses on
analyzing online texts  from sources such as Wikipedia, JW.org, Crúbadán \cite{scannell12007crubadan} and PanLex~\cite{kamholz-etal-2014-panlex}. They then extend their analysis to projects like Unilex,\footnote{https://github.com/unicode-org/unilex} CorpusCrawler~\cite{brawer2017corpus}, Bible.is and the LTI corpus for LangID~\cite{brown-2014-non}. Relying on texts to determine the correct script for a language may not be a robust method, as texts collected online can be noisy or may lack accurate labels.
We further discuss \citet{van-esch-etal-2022-writing}'s work
and its limitations in \secref{agreement}, including its
tendency to include the Latin script for languages for which romanization is not widely used.

\subsection{Applications}

\textbf{Corpus cleaning.}
 One of the uses of script identification is corpus cleaning. \citet{imanigooghari-etal-2023-glot500-acl} detect the script for each sentence and treat each language-script as a separate entity. They exclude all corpora for which the language-scripts are found to be incorrect or noisy; for example, when there is a mismatch between language and script, the corpus is removed.

\citet{kreutzer2022quality}
reported in their manual audit on multilingual datasets that
languages written in scripts other than their correct ones,
or text mixed with non-linguistic material, are good
indicators of a corpus being of low quality.

\textbf{Analysis of pre-trained models.} \citet{Acs2019}
studies the mBERT~\cite{devlin-etal-2019-bert} tokenizer vocabulary. They compile unicode ranges into 18 categories and use these ranges with regex to detect the script of vocabulary tokens.
Similar to \citet{Acs2019},
\citet{van-esch-etal-2022-writing}  analyzes the vocabulary
coverage of three models:
mBERT~\cite{devlin-etal-2019-bert},
XLM-R~\cite{conneau-etal-2020-unsupervised} and
mT5~\cite{xue-etal-2021-mt5}. Our analysis covers a wider
range of languages models and benefits from the alternative
methodology we adopt for \genericname.

\textbf{Fonts and keyboards.} One application of studying writing systems is the development of Unicode fonts, such as SIL Fonts,\footnote{https://software.sil.org/fonts/} or keyboards for  devices such as Keyman.\footnote{https://keyman.com/}

\textbf{OCR post-correction.}
Studying writing systems also finds practical use in improving OCR (optical character recognition) systems for rare languages. Take, for instance, the case where a Cyrillic character is erroneously replaced with a visually similar Latin letter. This highlights the importance of integrating post-correction methods into OCR systems~\citep{rijhwani-etal-2020-ocr}. A simple method to detect such errors would involve script identification performed on any part of the text to determine the percentage of characters in that part belonging to the same script.

\section{\dataname}
We now describe \dataname, a resource providing writing
system metadata for more than \numberlanguages language
varieties. We identify languages based on ISO~639.\footnote{https://iso639-3.sil.org/}

\subsection{Source selection}
We conducted an
exhaustive search to identify potential sources of
information about writing systems that we could use for our
goals in creating \genericname.
We focused on sources known for collaborative
contributions or recognized for their reliability across a
wide range of languages. We summarize the result of this
search as follows.

(i) \textbf{LREC\_2800 metadata}. \citet{van-esch-etal-2022-writing}
compiled a database containing information on
the writing systems of
more than 2,800 languages
under the CC BY 4.0 license, permitting modification and
redistribution.\footnote{https://github.com/google-research/url-nlp}

(ii) \textbf{Wikipedia metadata}. Wikipedia hosts pages for each
language's ISO 639 code and some of these pages include
details about the language's writing system. It also contains
information about writing systems that have not yet been
incorporated into Unicode. However, not all pages contain
metadata for writing systems. This dataset is  available
under the CC BY-SA 4.0 license, permitting modification
and
redistribution.\footnote{https://en.wikipedia.org/wiki/ISO\_639:\{ISO639\}}

(iii) \textbf{ScriptSource metadata.} Developed by SIL, ScriptSource
is a dynamic collaborative website serving as a reference
for writing systems. It gives information on which
languages use which script.
This dataset is available under the CC BY-SA 3.0 license, permitting modification and
redistribution.\footnote{https://scriptsource.org/scr/\{ISO15924\}}

(iv) \textbf{LangTag metadata.} The WSTech team of SIL offers writing
system metadata for language varieties
(format:  JSON).
This dataset is available under the MIT License,
permitting modification and
redistribution.\footnote{https://github.com/silnrsi/langtags}

(v) \textbf{Other sources.} We came across additional
sources during our search, but they have limited coverage
of languages. For example, the IANA language subtag registry
provides script metadata for 134
languages.\footnote{https://iana.org/assignments/language-subtag-registry}
Other sources are consulted by sources (i) -- (iv),
for example,
Omniglot\footnote{https://www.omniglot.com/writing/langalph.htm}
in LREC\_2800 and Unicode
CLDR\footnote{https://github.com/unicode-org/cldr-json} in
LangTag.

Note that none of these sources cover all languages, and
there is a potential for some languages to have incorrect
scripts listed (see \secref{agreement}).  To address this,
we  incorporate all four sources (i) -- (iv) in \dataname;
this allows us to give preference to script identification decisions that several
sources agree upon.
We gathered the Wikipedia and
ScriptSource data -- which
are not accessible in tabular format -- by crawling.

\subsection{Preprocessing}
There is a total of 8030 unique three-letter ISO 639 codes that at least one of the sources covers. The most used version of the ISO~639 code set in
the NLP community is ISO~639-3; however, not all 
three-letter codes are part of this subset. For instance,
ber (Berber languages) is part of code sets 639-2
and 639-5, but not part of 639-3. To handle this, we include all 
three-letter ISO codes, not just those from ISO~639-3.

The number of three-letter ISO~639 codes covered is
2836 for LREC\_2800, 1726 for Wikipedia, 7875 for ScriptSource and
7901 for LangTag.

\subsection{Agreement}\seclabel{agreement}

We assess agreement between two metadata
sources using Jaccard similarity:
\[
J(A, B) = \frac{|A \cap B|}{|A \cup B|}
\]
where \(A\) and \(B\) are sets of scripts given for an ISO
code by the two sources in question.
Since 
the Wikipedia data is of a smaller size and does not
represent writing systems in a uniform format
(e.g., ISO~15924), we use Wikipedia as a secondary source
of information when merging information, especially in cases
where there is no agreement.

\begin{table}[thp]
\centering
\resizebox{0.99\linewidth}{!}{
    \begin{tabular}{l|c|c|c|r}
        \textbf{Pair} & \textbf{$|\mathcal{L}|$} & \textbf{CA} & \textbf{PA} & \textbf{NA} \\
        \midrule
        (LangTag, LREC\_2800) & 2814 & 2385 & 404 & 25 \\
        (ScriptSource, LREC\_2800) & 2811 & 2372 & 414 & 25 \\
        (LangTag, ScriptSource) & 7858 & 7567 & 287 & 4 \\
    \end{tabular}
}
\caption{Agreement counts for each  pair of sources. CA: complete agreement ($J = 1.0$), PA: partial agreement ($0 < J < 1$), NA: no agreement ($J = 0$), $|\mathcal{L}|$: number of common ISO 639 codes.}

\tablabel{agreementstats}
\end{table}

We present the results for each pair in LangTag,
ScriptSource and LREC\_2800 in \tabref{agreementstats}. 
LangTag and ScriptSource completely agree
($J =1.0$) for 96\% of ISO codes. This is not surprising,
given that both sources are from SIL. However,
some disagreements still exist. Additionally, it appears
that LangTag aligns more closely with LREC\_2800, as it
shares a greater number of ISO 639 codes,
fewer partial agreements ($0 < J < 1$) and no disagreements ($J
= 0$).

To understand the discrepancies between different metadata
sources, we conducted a manual analysis, and observed the following trends.\footnote{We analyse discrepancies for languages which Wikipedia data knowledge is available. Our approach involves relying on information sourced from Wikipedia and other web-based resources for each language.} 

(i) \textbf{Rare or historic scripts.} ScriptSource and LangTag metadata tend to include rare and historic scripts. For instance, in the case of Turkish (tur), alongside the primary Latin script, these sources also list Arabic, Greek and Cyrillic. In contrast, LREC\_2800 exclusively lists Latin, the current official script.

(ii) \textbf{Romanized versions.} LREC\_2800 often
introduces a Latin version for a language, even if it is
rarely used. For instance, there is a Latin entry for fas
(Farsi) in LREC\_2800, despite it not being the official
script and not widely used, even in social
networks. Scriptsource and langtag only report
Arabic.\footnote{Ultimately, the meaning of ``widely used''
depends on the intended application. If
data entered on smartphones is the primary interest of a
company, then Latin may actually be not infrequent because
text messages may be entered in the Latin script, for example,
by Farsi speakers in countries where the Latin script dominates.}

(iv) \textbf{Partial information.} There are instances where
each source only partially supports certain scripts. For
example, ScriptSource and LangTag mention that aat (Arvanitika
Albanian) uses the Greek script while LREC\_2800 reports the use of  
the Latin script. However, resolving this conflict is difficult because there is no proper definition. There is disagreement among Arvanitika speakers about using the Greek versus the Latin script. Additionally, this language is classified as an endangered language. The writing history for this language is outlined in \citet{sasse1991arvanitika}.

(v) \textbf{Errors.}
There are instances where it is clear that a language is
highly unlikely to be written in a particular script. One of
these cases is var (Huarijio) in LREC\_{2800},
which is indicated to be written in Devanagari.
In our judgement, this is an error since
var is a Uto-Aztecan language spoken in northwestern Mexico and
Devanagari is only used for South Asia languages.

\subsection{Compilation}
\seclabel{compilation}

We now explain the process of compiling \dataname by
combining different metadata sources. As will be apparent
from our discussion, creating a reliable
writing system resource is not straightforward.

Two of our main desiderata are \emph{usefulness for NLP} and
\emph{accuracy}.

As far as usefulness for NLP is concerned,
if we accepted all scripts that any of the sources lists 
for a language, then
we would include errors and scripts that in practical NLP
contexts are very unlikely to be relevant.
The most important instance of this is that some sources give
Latin as a valid script in many cases where its use is
extremely rare. Including Latin for such languages would be
harmful for use cases such as corpus quality control. For
example, a non-Farsi subcorpus written in Latin cannot be
excluded using script identification if we accept Latin as a standard script for Farsi.

On the other hand,
the desideratum of accuracy demands that we do not simply
adopt a criterion of perfect agreement of the four
sources. Such a heuristic would
exclude important language metadata that might be useful to the
NLP community.

To allow users of \toolname to trade off usefulness against accuracy,
we define two metadata categories:
\MAIN and AUXILIARY.
The \MAIN metadata give the primary scripts based on consensus
among the metadata sources. The AUXILIARY metadata
give secondary scripts, those that are only specified as admissible
by a single source.

Given the 96\% complete agreement between LangTag and
ScriptSource, we prioritize resolving disagreements between
these two sources using information from Wikipedia and
LREC\_2800. We merge LangTag and ScriptSource as one
consolidated group named SIL, which is the aggregation of
LangTag and ScriptSource if they match or if the
discrepancies can be resolved based on additional
resources. Only those
discrepancies that cannot be resolved this way are collected
in SIL2-aux.

As a result of this consolidation, we have now three
metadata sources: SIL, LREC\_2800 and
Wikipedia.
Given a language $l$ identified by an ISO 639 code,
we categorize a script for $l$ as \MAIN if this is
supported by at least two of the three sources
(e.g., the \MAIN metadata specify
Kpel as one of the admissible scripts
for kpe (Kpelle)
since SIL and Wikipedia agree on it even though
LREC\_2800 does not)
or if only one of three sources provides information about
admissible scripts for $l$.

(i) In cases of partial information, such as for
aat (Arvanitika Albanian), where both LREC\_2800 and
Wikipedia agree on Latin, and both Wikipedia and SIL agree
on Greek, we include both Latin and Greek in \MAIN.

(ii) If only one metadata source reports a script and not
the others, the script is placed in the auxiliary category
specific to that source. Wiki-aux,
LREC2800-aux, and SIL-aux are used for Wikipedia,
LREC\_2800, and SIL, respectively. SIL2-aux is exclusively
used for discrepancies between ScriptSource and LangTag.

\section{\toolname}
We now describe \toolname, an open-source Python tool
that identifies the writing systems of input text. It
supports the 161 scripts
in Unicode 15.0,
identified as ISO 15924 codes. \toolname is the first tool to
provide labels based on ISO 15924 with this level of coverage. \figref{toolexample} gives an example of how to  use \toolname.

\begin{figure}
    \centering
    \includegraphics[width=0.49\textwidth]{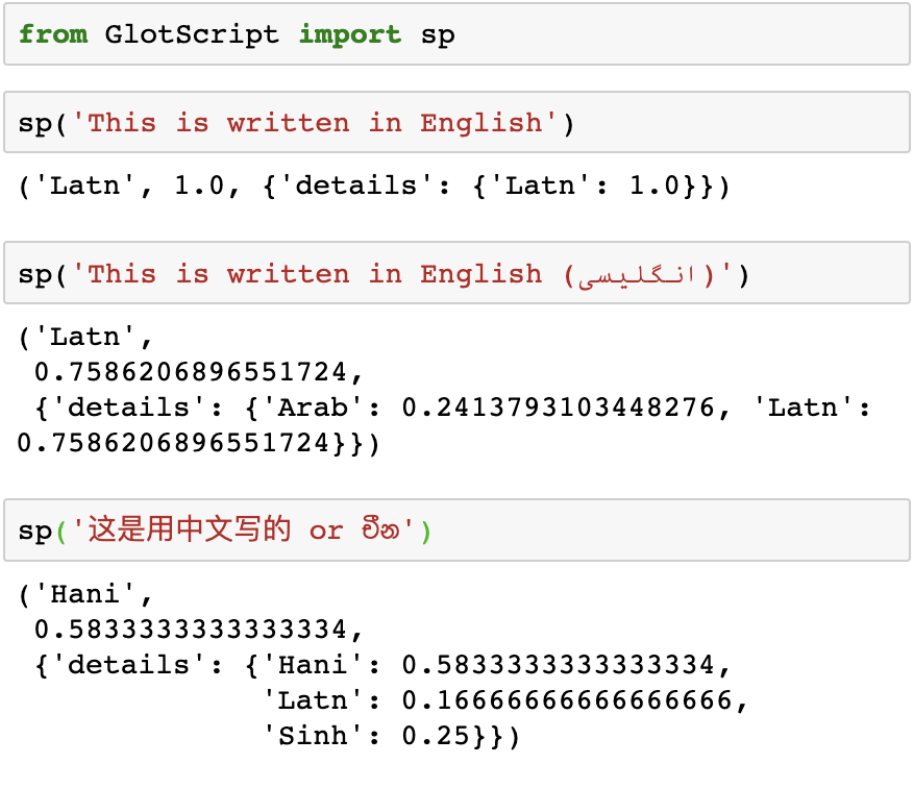}
\caption{\figlabel{toolexample} How to use \toolname: three
    examples.
\toolname returns a tuple consisting of the main script, the percentage of characters in the main script and detailed information on the distribution of scripts.}
\end{figure}

\subsection{Development}
We first sorted unicode ranges into different script
categories,
based on the Unicode Character
Database.\footnote{https://unicode.org/Public/15.0.0/ucd/Scripts.txt}
We then matched these ranges with ISO 15924 code names from
Wikipedia.\footnote{https://en.wikipedia.org/wiki/ISO\_15924}

For an input text,
\toolname identifies
the unicode range of each character, maps it to an ISO 15924
code and then calculates the percentage of each script.
\toolname returns the main script (the one that 
most characters belong to) and detailed information on the
distribution of scripts.

\subsubsection{Special codes} 
\toolname also uses three special codes that are not proper scripts.

(i) \textbf{Zzzz.} This code is used for unknown Unicode ranges. We also add the replacement character (U+FFFD \ucr) as Zzzz.
    
(ii) \textbf{Zinh.}
This code is assigned to a character who inherits its
script from the previous character. 
 For example, the zero width joiner character
    (U+200D) is used for joining characters. It  does not
    belong to any script, but rather inherits its script
    code from the immediately preceding character.
    
(iii) \textbf{Zyyy.}
This is the ISO 15924 code for undetermined script. This
script code covers
characters like punctuation, symbols, mathematical notation and
musical notation that are used across many different scripts.

\subsubsection{Efficiency} We randomly generate a test set
of 1 million sentences, each with a length of 100, using
characters from different Unicode ranges.  The walltime of
processing this test set with \toolname on a single core of
an Intel Xeon E7-8857 3GHz CPU is 80.790 seconds, i.e.,
about $8 \times 10^{-5}$ seconds per sentence.

\section{Experimental setup}

We present experiments for two tasks to demonstrate the
usefulness of \toolname and \dataname.

(i) \textbf{Corpus quality assessment.} We investigate
multilingual datasets by determining if a text assigned by
the corpus metadata to
a particular language
is written in a script that is
admissible for that language.
If this is not the case for a particular text, it hints at
mislabeling and suggests that the text most likely  belongs to another
language or is noise. This part of our experiments highlights
the benefits of a script identification tool
for creating high-quality corpora for low resource languages.

(ii) \textbf{Multilingual models.} We quantify the presence
of each script within the vocabulary of several multilingual
language models, focusing on large multilingual language
models. We evaluate the level of representation of each
script, which sheds light on the quality of representations
of languages using that script.

\subsection{Corpus quality assessment}
\dataname{} lists for each language $l$, identified by an ISO~639 code,
the scripts that are
commonly used for $l$.
Recall that, as shown in \figref{toolexample}, 
the function $sp(s)$ provided by \toolname computes the
percentage of each script
in the input and identifies the main script.

Let $s$ be an input sentence from a corpus that is assigned
the ISO~639 code $l$ by the corpus metadata.
The predicted main script for  $s$ -- i.e., $sp(s)$ --
is either one of the admissible scripts (according
to \dataname) for $l$. We call this a match. Or it is not
one of the admissible scripts. We call this a mismatch. In
case we find a mismatch for $s$,
we evaluate this as an error. We refer to this heuristic as
the \textbf{script mismatch rule}.
We determine for each sentence of the corpus whether it is a
match or a mismatch and then report the proportion of errors.

\input{supplementary/mc4-oscar}

\subsubsection{Evaluation corpora}
We select two corpora that have been recognized in multiple past studies
for their multilinguality and the inclusion of lower resource languages.

(i) Multilingual C4 (mC4)~\cite{xue-etal-2021-mt5} is a
document-level dataset used for training the mT5 language
model. It uses CLD3~\cite{botha-etal-2017-natural,
salcianu2018compact} language identification (LID).  CLD3
supports 107 languages. Accordingly, mC4 contains monolingual
texts in 107 languages.
    
(ii) OSCAR~22.01~\cite{OrtizSuarezSagotRomary2019,
AbadjiOrtizSuarezRomaryetal.2021} is a set of monolingual
corpora for 151 languages. It is deduplicated and uses
Fasttext~\cite{joulin2017bag}
FT176\footnote{https://fasttext.cc/docs/en/language-identification.html}
LID on a line by line level.

Both corpora are sourced from CommonCrawl.
\citet{kreutzer2022quality} performed a manual audit  on  100
sentences (or less) per language for these two corpora.

\subsubsection{Setup}
We load both datasets using the Hugging Face API.\footnote{https://huggingface.co/datasets} Each row
in the dataset is split by \texttt{\textbackslash n} (which
we consider to be the sentence delimiter) and deduplicated.

For both corpora, we randomly select 1000 sentences per language.
We exclude languages for which there are fewer than 1000
sentences available, resulting in a coverage of 118 languages from
OSCAR~22.01. For example, for dsb (Lower Sorbian), diq
(Dimli/Zaza) and eml (Emiliano-Romagnolo), there is
only one sentence each
available
in OSCAR~22.01, so we exclude these languages. We map the language identifiers provided by
the corpus metadata to three-letter ISO~639 codes.

We apply \toolname
to the 1000-sentence subsets per language
and obtain the main script for each
sentence. We apply the script mismatch rule to identify the
sentence as correct or incorrect. However, if the corpus
metadata specify the script in addition to the language
(e.g., bg-Latn), then we only consider the script given as a
candidate script for that sentence by the metadata (e.g., we
only consider Latn for bg-Latn).

\subsection{Multilingual models}

We analyze the representation of common writing systems in state-of-the-art pretrained models.
Most of these models are claimed to be highly multilingual. We
approach this analysis employing the following two methods.

(i) Following \cite{van-esch-etal-2022-writing, Acs2019},
we examine the writing systems present in the vocabulary of each model's tokenizer.

(ii) We tokenize the multi-parallel corpora
of \udhr\footnote{http://unicode.org/udhr/d/} using each
model's tokenizer.
For each writing system, 
we then measure the number of tokens
generated and the percentage
of unknown tokens (UNK) generated.  Similar experiments are also conducted by \citet{ahia-etal-2023-languages} and \citet{petrov2024language} on FLORES-200~\citep{flores-101, nllbteam2022language}. \udhr dataset supports a greater variety of languages and scripts compared to FLORES-200.

\subsubsection{Model selection} We select ten state-of-the-art models for 
their multilingual capabilities or for their frequent use:
GPT-4~\cite{openai2023gpt4},
Falcon~\cite{refinedweb},
Llama 2~\cite{touvron2023llama},
BLOOM~\cite{scao2022bloom},
Glot500~\cite{imanigooghari-etal-2023-glot500-acl},
XLM-R~\cite{conneau-etal-2020-unsupervised},
mBERT and BERT~\cite{devlin-etal-2019-bert},
mT5~\cite{xue-etal-2021-mt5} and
NLLB~\cite{nllbteam2022language}.

\subsubsection{\udhr} UDHR consists of more than 500 translations of
the Universal Declaration of Human Rights, each containing
30 short articles.
We remove all the translations that are incomplete (fewer than
89 sentences) or noisy
(e.g., lines consisting of the single English word `missing').
We ensure that all 30 articles are available in a
translation and that it has a valid ISO 639-3 code (not
undetermined). In cases where multiple versions are
available for a pair of ISO 639-3 and ISO 15924, we make a
random selection.
This procedure selects a subset of UDHR that covers 396
different language-scripts.

\section{Results and analysis}\seclabel{results}

\subsection{Corpus quality assessment}

\tabref{mc4-oscar}
reports the five top and bottom
languages (in terms of inferred accuracy of their metadata)
for each corpus,
along with
correct and incorrect scripts. Scripts
highlighted in \greenund{green} are deemed correct based
on \dataname \MAIN. \yellow{yellow} represents the
scripts that were returned for AUXILIARY. Note that quite
a few languages have 
Latin as an AUXILIARY script, based on the
LREC\_2800 metadata. The ACC column
displays the accuracy of the correct script based only on
\MAIN.

For the 118 selected languages in OSCAR, we obtain an
average script accuracy of 0.947. For the 107
languages in mC4, the average score is 0.917. These averages
are high, indicating a favorable quality overall. However,
when examining the bottom five languages with the lowest
correct script scores, the average drops to 0.823 for OSCAR
and 0.566 for mC4. 

Based on our audit of common errors in the OSCAR corpus, we
can confirm that incorrect Latin sentences are either written in
English or are related to website content, such as website
functionalities (comment and search sections), URLs and
dates. This confirms that including more scripts,
especially all the romanized versions, in the writing
metadata even when their use is not solidly attested would hamper our
ability to identify incorrect sentences in low resource
corpora.
This is why we decided that
when merging different datasets, if a script is not approved
by the majority of sources, it will be kept in  AUXILIARY
 (see \secref{compilation}). We also noticed that most sentences with script
mismatches are short. We therefore run another set
of experiments, this time using a length-based filter that
keeps either
70\% (ACC70 column) or 50\% (ACC50 column) of the longest sentences.

For the bottom languages of OSCAR in \tabref{mc4-oscar}, it is
clear that length filtering proves to be
effective. Notably, for amh (Amharic), the accuracy improves
0.118 when retaining only the 50\% longest
sentences. However, this is not the case for the bottom
languages of mC4, particularly for cym (Welsh)
and snd (Sindhi) where the accuracy
worsens. Additionally, the correct scripts for these two
languages are not the most frequent in their respective
corpora. This suggests that the mistakes are not merely
short incorrect sentences, but rather lengthy paragraphs in
the wrong language. In the case of Welsh, upon closer
inspection, it becomes apparent that the incorrectly
identified Greek scripts are actually written in ell (Modern
Greek). We also observed suspicious patterns in the
Latin portion of this data but it also contained many correctly
written sentences in cym (Welsh). For snd (Sindhi), the data
contains numerous extensive paraphrases in English, and we
also suspect a mix of ara (Arabic) and fas (Farsi) in the
Arabic script part.

The infrequent instances of incorrect writing systems in
OSCAR may indicate the effectiveness of line-level LID
filtering. These results hint at the need for further
research on LID. Additionally, we recommend that in
newly published LID and corpora, along with the language
code, a script code should be assigned to each sentence as
part of the metadata. Adopting this recommendation would
significantly facilitate error prevention.

\begin{figure}[htp]
    \centering
    \includegraphics[width=0.49\textwidth]{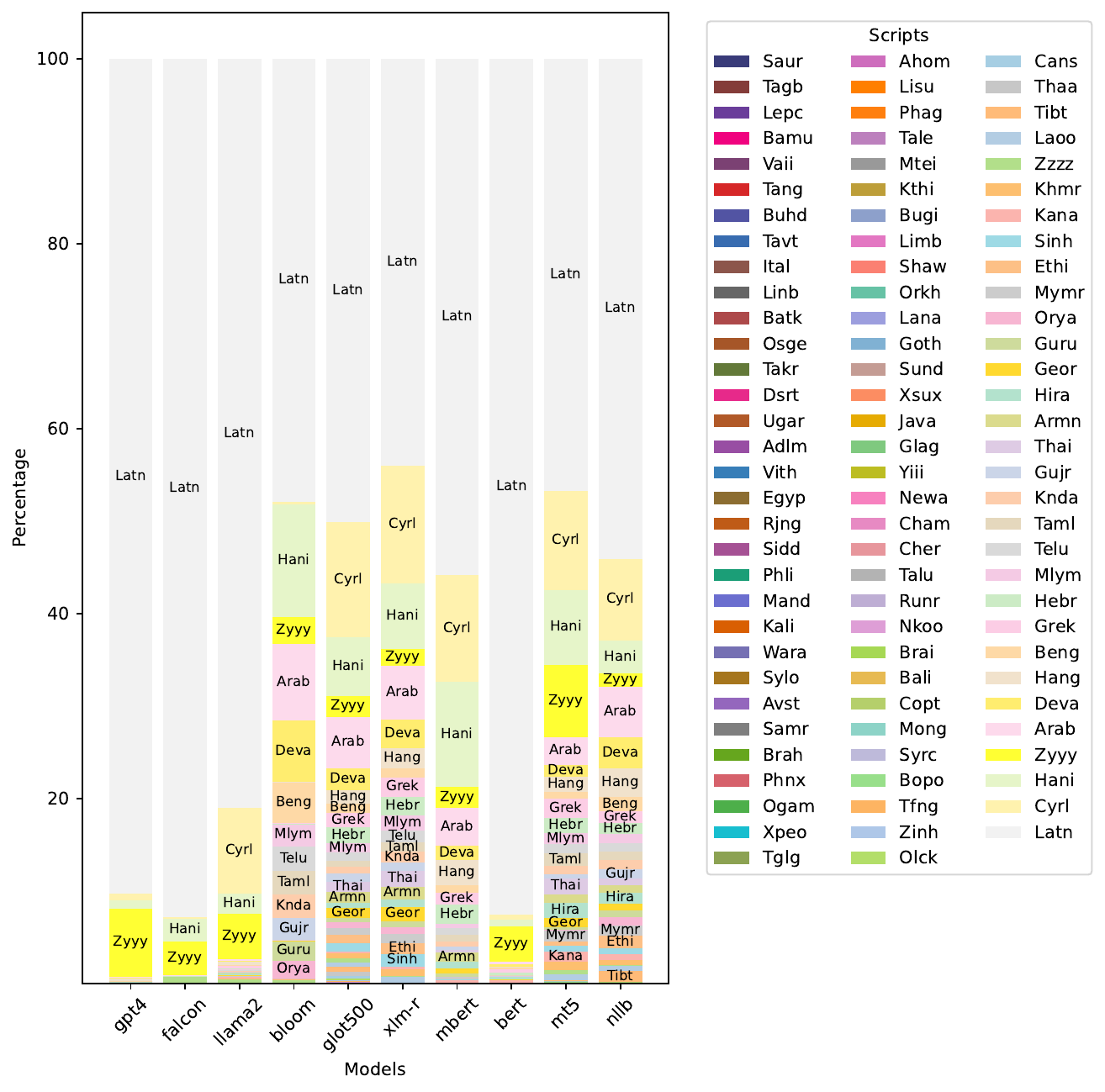}
    \caption{The percentage of each script in the vocabulary of model tokenizers. Scripts with a presence of more than 1\% in each tokenizer are text-labeled in the figure.
}
    \figlabel{tokenizer-vocab}
\end{figure}

\begin{figure*}[t]
    \centering
    \includegraphics[width=0.99\textwidth]{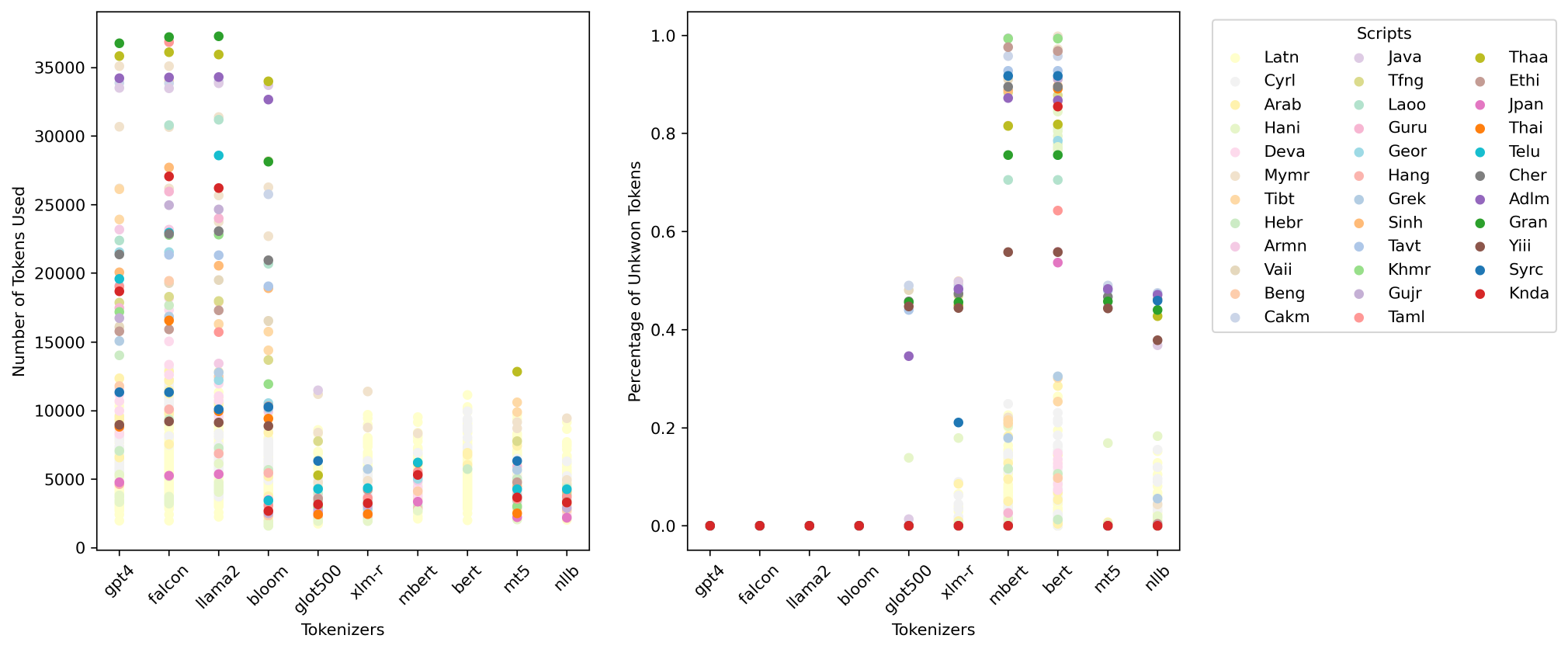}
    \caption{Analysis of the
      multilinguality of the tokenization of ten language
      models. This analysis was performed on 396 UDHR translations.
      Left: the number of tokens into which the UDHR
      translation is tokenized. We omit a pair of tokenizer and translation with more than 5\% unknown tokens.
Right: the percentage of unknown tokens generated for a pair
of tokenizer and translation.\protect}
    \figlabel{tokenizer-udhr}
\end{figure*}

\subsection{Multilingual models}

\subsubsection{Tokenizer vocabulary}

We use \toolname to analyze the token
vocabulary of the ten language models and determine each token's
script.
\figref{tokenizer-vocab} reports the
percentage distribution of each script for each tokenizer's
vocabulary.

(i) The Cyrillic representation in the BLOOM tokenizer is relatively scarce compared to other models.

(ii) The BERT tokenizer supports not only Latin scripts but also recognizes Hani, Arabic, Cyrillic and some tokens in an additional 12 scripts.

(iii) Glot500 encompasses the highest number of scripts,
  totaling 88, followed by mT5, which supports 66. However,
  a significant portion of these scripts in both models has limited presence. 

(iv) Llama2's second most prominent script is Cyrillic.

(v) Falcon's second most prominent script is Hani.

(vi) The GPT-4 tokenizer vocabulary includes
representations for 18 scripts, albeit not very
comprehensively compared to its coverage of Latin.

(vii) In all tokenizer models combined, a total of 92 scripts has some presence.

\subsubsection{UDHR tokenization}

Parsing the UDHR translations with
the specific tokenizer associated with each model,
we
generate a plot illustrating the token count required by
each model to tokenize the UDHR translation. Since not all
model tokenizers operate at the byte level, this may result
in the generation of unknown (UNK) tokens. We only consider
tokenizer-translation pairs where fewer than 5\% unknown
tokens are produced. \figref{tokenizer-udhr} displays 
the token count used by each tokenizer (left) and the
percentage of unknown tokens (right). Rather than coloring the plot
data based on language labels, we choose to use script
categories for color representation.

GPT-4, in addition to being trained on
English, was also trained on some other
languages. For instance, it is capable of translating
between English and sin (Sinhala).  In tasks such as text
generation, the number of generated tokens is particularly
important. For example, for the English UDHR translation, the
GPT-4 tokenizer
produces 1983 tokens. However, for the Sinhala UDHR
translation, it generates 20,071 tokens, nearly 10
times more. As the
pricing of OpenAI APIs is also based on the number of
tokens, this demonstrates that generation of Sinhala is very
expensive using GPT-4 in comparison with English.

\section{Conclusion}

We publish \dataname, an extensive resource covering
writing systems for over \numberlanguages languages,
including thousands of typically overlooked ``lowest-resource''
languages. We open-source \toolname, a script
identification tool that supports all 161 scripts in Unicode
15.0. It reports the script distribution within a given
text, using ISO 15924 labels. This work is the first to
create a highly efficient tool for script identification and
output labels based on ISO 15924 with this level of
coverage.

We apply \dataname and \toolname to the task of corpus quality assessment. Our findings indicate that these two components work together effectively to improve the quality of existing low resource corpora.
Furthermore, we investigate the tokenizers of large language models like GPT-4. This analysis enables us to assess how well a script is represented, serving as an indicator of the representation quality of languages written in that script.

In future work, we aim to expand \dataname by
offering a better categorization of writing systems
such as "live", "rare", "historic",
"romanization present", "romanization in use". We also would like
to include more metadata.

Based on the lessons we learned from conducting this study,
we recommend that creators of
language identification tools and
text corpora provide,
along with the language code, a script code
for each sentence as part of the
metadata.
Adopting this recommendation could be of great benefit for
error prevention and better quality of language resources.

\section*{Limitations}
We acknowledge that some of the input metadata may contain errors; we rely on consensus to decrease this risk. However, there is a potential risk of excluding a writing system for a language or including a noisy one during the collection and processing of NLP corpora.

\section*{Ethics statement}
The resources used in our study come from openly available metadata with permission for modification and redistribution.
We  make our research artifacts,
\dataname and \toolname openly available
to foster collaboration and reproducibility.

\section*{Acknowledgements}
We would like to thank anonymous reviewers. This
work was funded by the European Research Council (grant \#740516).

\bibliography{main}

\end{document}

%% file: supplementary/mc4-oscar.tex
\begin{table*}[thp]
\centering
\resizebox{\linewidth}{!}{
    \begin{tabular}{l| l | l |l |c|c|c}
        & & \textbf{Corpus Code:}  &  & &  &\\
        & & \textbf{ISO 639-3}  & \textbf{Scripts} & \textbf{ACC$\uparrow$}  & \textbf{ACC70$\uparrow$} & \textbf{ACC50$\uparrow$}\\
        \midrule
        \midrule
        \multirow{5}{*}{\rotatebox{90}{Highest ACC}} &
        \multirow{11}{*}{\rotatebox{90}{mC4} }
        
        & st:sot (S Sotho)  & \greenund{Latn}:1000 & \textbf{1.000} & \textbf{1.000} & \textbf{1.000}\\
        & & fil:fil (Filipino) & \greenund{Latn}:998, Cyrl:1, Hani:1	 & 0.998 & 0.999 & \textbf{1.000}\\
        & & ro:ron (Romanian) & \greenund{Latn}:996, Zyyy:4, \greenund{Cyrl}:1	& 0.995 & 0.997 & \textbf{1.000} \\
        & & id:ind (Indonesian) & \greenund{Latn}:995, Zyyy:3, Hani:1, Hebr:1	 & 0.995 & \textbf{1.000} & \textbf{1.000} \\
        & & sw:swa (Swahili)  & \greenund{Latn}:995, Zyyy:5	 & 0.995 & \textbf{1.000} & \textbf{1.000} \\
        \cline{1-1} \cline{3-7} \rule{0pt}{2.5ex}
        \multirow{5}{*}{\rotatebox{90}{Lowest ACC}}

        & & ne:nep (Nepali) & \greenund{Deva}:609,
        Hani:219, Latn:88, Hang:44, Thai:12,
        Laoo:8, Zyyy:8, Orya:7, Other:5 
        & 0.609 & 0.730 & \textbf{0.797}\\

        & & mn:mon (Mongolian) & \greenund{Cyrl}:502, Hebr:348, \yellow{Latn}:135, Zyyy:14, Hani:1 & 0.502 & 0.557 & \textbf{0.570}\\
        
        & & cy:cym (Welsh) & Grek:603, \greenund{Latn}:367, Zyyy:11, Hebr:9, Cyrl:5, Zzzz:4, Arab:1 & \textbf{0.367} & 0.338 & 0.295\\

        & & sd:snd (Sindhi) & \yellow{Latn}:654, \greenund{Arab}:329, Zyyy:12, Zzzz:2, Cyrl:1, Hang:1, Telu:1	& \textbf{0.329} & 0.271 & 0.222 \\
        
        & & mr:mar (Marathi) & Hani:454, Thai:252,
        \yellow{Latn}:119, \greenund{Deva}:116,
        Zyyy:34, Guru:10, Beng:4, Khmr:3, Other: 8
        & 0.116 & 0.136 & \textbf{0.141} \\

        \midrule
        \midrule
        \multirow{5}{*}{\rotatebox{90}{Highest ACC}} &
        \multirow{10}{*}{\rotatebox{90}{OSCAR} }
        & id:ind (Indonesian) & \greenund{Latn}:998, {Zyyy}:2 & 0.998 & \textbf{1.000} & \textbf{1.000}\\
        & & war:war (Waray) & \greenund{Latn}:997, {Zyyy}:3 & \textbf{0.997}  & \textbf{0.997} & 0.996 \\
        & & als:gsw (Swiss G) & \green{Latn}:996, {Zyyy}:3, {Cyrl}:1 & 0.996 & 0.996 & \textbf{1.000}\\
        & & vo:vol (Volapük)  & \greenund{Latn}:994, {Arab}:4, {Cyrl}:1 & 0.994 & \textbf{1.000} & \textbf{1.000}\\
        
        & & nds:nds (Low G) & \greenund{Latn}:994, Zyyy:2, Cyrl:2, Hang:1, Thaa:1 & 0.994 & \textbf{1.000} & \textbf{1.000}	\\
        
        \cline{1-1} \cline{3-7} \rule{0pt}{2.5ex}
        \multirow{5}{*}{\rotatebox{90}{Lowest ACC}} 

       && am:amh (Amharic) & \greenund{Ethi}:822, \yellow{Latn}:164, Zyyy:12, Hani:1, \yellow{Arab}:1	&  0.822 & 0.883 & \textbf{0.940}\\

        && gu:guj (Gujarati) & \greenund{Gujr}:802, \yellow{Latn}:180, Zyyy:12, Deva:6	 & 0.802 & 0.863 & \textbf{0.883}\\

        && si:sin (Sinhala) & \greenund{Sinh}:801, \yellow{Latn}:188, Zyyy:11 & 0.801  &  0.905 & \textbf{0.948}\\

        & & th:tha (Thai) & \greenund{Thai}:800, \yellow{Latn}:181, Zyyy:18, Hani:1	& 0.800 & 0.883 & \textbf{0.917}\\
        
        & & te:tel (Telugu) & \greenund{Telu}:799, \yellow{Latn}:188, Zyyy:9, Deva:3, Cyrl:1 & 0.799 &  0.880 & \textbf{0.908}\\

    \end{tabular}
}
\caption{Script accuracy for mC4 (top) and OSCAR (bottom) corpora.
We display the five best-performing and worst-performing languages.
\greenund{Green} indicates correct scripts based on
\dataname \MAIN. \yellow{Yellow} indicates correct scripts
based on \dataname AUXILIARY. ACC:
accuracy, i.e., the proportion of sentences for which the
script identified by \toolname is one of the admissible
scripts (according to \dataname) of the language provided by
corpus metadata for the sentence.
ACC70/ACC50: accuracy for the 70\%/50\%  longest
sentences. To save space, we write "Other" for multiple
scripts with a small number of sentences. The best scores
are bolded for each row. S Sotho = Southern Sotho. Swiss/Low
G = Swiss/Low German.}
\tablabel{mc4-oscar}
\end{table*}